\documentclass[runningheads]{llncs}
\usepackage{xcolor,soul,framed} 

\colorlet{shadecolor}{yellow}
\usepackage[pdftex]{graphicx}
\usepackage{caption}
\usepackage{subcaption}
\usepackage{tabu}
\graphicspath{ {./images/} }
\DeclareGraphicsExtensions{.pdf,.jpeg,.png}

\usepackage[cmex10]{amsmath}
\usepackage{array}
\usepackage{mdwmath}
\usepackage{mdwtab}
\usepackage{eqparbox}
\usepackage{url}

\usepackage{floatrow}
\newfloatcommand{capbtabbox}{table}[][\FBwidth]

\usepackage{blindtext}

\usepackage{lipsum}
\usepackage{graphicx}
\usepackage{algorithm}
\usepackage{algpseudocode}
\usepackage{amsmath}
\usepackage{diagbox}
\usepackage{makecell}
\usepackage{subcaption}
\usepackage{amssymb} 
\usepackage{enumitem}
\usepackage{multirow}
\usepackage{xspace}
\newcommand\Tstrut{\rule{0pt}{2.6ex}}         
\newcommand\Bstrut{\rule[-0.9ex]{0pt}{0pt}}

\usepackage{todonotes}

\makeatletter
\DeclareRobustCommand\onedot{\futurelet\@let@token\bmv@onedotaux}
\def\bmv@onedotaux{\ifx\@let@token.\else.\null\fi\xspace}
\def\eg{\emph{e.g}\onedot} 
\def\ie{\emph{i.e}\onedot}

\def\wrt{w.r.t\onedot} 
\def\etal{\emph{et al}\onedot}
\makeatother

\begin{document}
\title{Relation Embedding for Personalised Translation-based POI Recommendation}
\titlerunning{Relation Embedding for Personalised POI Recommendation}
\author{Xianjing Wang\inst{1,2} \and
Flora D. Salim\inst{1,4}\orcidID{0000-0002-1237-1664} \and
Yongli Ren\inst{1}\orcidID{0000-0002-3137-9653} \and Piotr Koniusz\inst{2,3}\thanks{This paper is accepted by the PAKDD'20. 
}
}
\authorrunning{X. Wang et al.}
%
\institute{RMIT University, Melbourne, Australia \\
\email{\{xianjing.wang, flora.salim, yongli.ren\}@rmit.edu.au} \and
Data61/CSIRO, Canberra, Australia\\
\email{piotr.koniusz@data61.csiro.au} \and
Australian National University, Canberra, Australia \and
Corresponding author
}


\maketitle

\begin{abstract}

Point-of-Interest (POI) recommendation is one of the most important location-based services helping people discover interesting venues or services. 
However, the extreme user-POI matrix sparsity and the varying spatio-temporal context pose challenges for POI systems, which affects the quality of POI recommendations. 
To this end, we propose a translation-based relation embedding 
for POI recommendation. 
Our approach encodes the temporal and geographic information, as well as semantic contents  effectively in a low-dimensional relation space by using Knowledge Graph Embedding techniques. 
To further alleviate the issue of user-POI matrix sparsity, a combined matrix factorization framework is built on a user-POI graph to enhance the inference of dynamic personal interests by exploiting the side-information. 
Experiments on two real-world datasets demonstrate the effectiveness of our proposed model. 



\end{abstract}
\section{Introduction}
With the increase of mobile devices on the market and ubiquitous presence of wireless communication networks, people gain easy access to Point-of-Interest (POI) recommendation services. A great number of Location-based Social Networks (LBSNs) have consequently been established \eg, Foursquare, Gowalla, Facebook Places, and Brightkite. 
The LBSNs often provide POI services that recommend users new POI venues that meet specific user criteria. 
In this paper, we develop a high quality personalized POI recommendation system by leveraging user check-in data. 
There are three technical challenges listed as follows:

\vspace{0.05cm}
\noindent{\textbf{Sparsity of user check-in data.}} One of the major challenges is to overcome the sparsity in the user check-in data. The user-POI matrix can be extremely sparse despite of millions of POIs and users in LBSNs. 

\vspace{0.05cm}
\noindent{\textbf{Temporal reasoning.}} Location-based POI recommendation systems utilize the temporal context 
\cite{xie2016learning}  
for the purpose of modeling personal preferences. 
The temporal information reflects users' needs and choices throughout the day. 

\vspace{0.05cm}
\noindent{\textbf{Spatial reasoning.}} A user's current geographical location limits their choice of check-in POIs \cite{liu2013personalized}. Many approaches model  relations between a user's current geographical location and their preferences with respect to the surrounding POIs. 
%

\begin{figure}
\vspace{-0.4cm}
\begin{floatrow}
\ffigbox{%
  \includegraphics[width=2.6in]{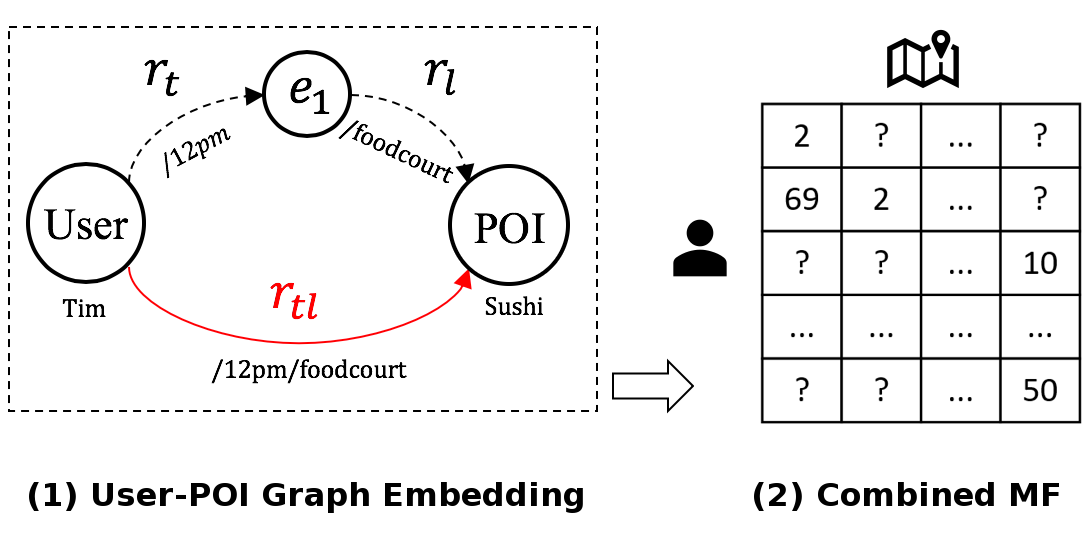}
}{%
  \caption{Overview of our GERec model.}\label{fg:overview}
}
\qquad
\capbtabbox{%
\vspace{-1cm}
  \begin{tabular}{cc}
\small
\centering 
\scalebox{0.75}{
\begin{tabular}{ c | c}
\hline 
\rule{0pt}{15pt} 
$h \stackrel{r}{\longrightarrow} t $
& $user \stackrel{12pm}{\longrightarrow} e_{1} \xrightarrow{food court} \textit{sushi shop}$  
\\ 
\hline 
$(h, r, t)$ & 
\makecell{$(u, r_{t}, e_{1})$ and $(e_{1}, r_{l}, v)$ \\ {\color{red} $(u, r_{t} \circ r_{l}, v)$} }
\\
\hline  
$\textbf{h} + \textbf{r} = \textbf{t}$ & \makecell{$\textbf{u} + \textbf{r}_t = \textbf{e}_1$ and $\textbf{e}_1 + \textbf{r}_l = \textbf{v}$ \\ {\color{red} $\textbf{u} + (\textbf{r}_{t} \circ \textbf{r}_{l}) = \textbf{v}$} }
\\
\hline
\end{tabular}}
  \end{tabular}
}{%
  \caption{Relation path embedding.}\label{tb:relation_path2}
}
\end{floatrow}
\end{figure}

\vspace{-0.4cm}

The above issues are often addressed by the use of side information in traditional recommendation systems. 
Such a side information may be retrieved from social networks \cite{lian2014geomf} 
and may include user demographic information, item attributes, and context information \cite{yin2016adapting}.
As the auxiliary data is useful for the recommendation systems \cite{shi2018heterogeneous}, it is desirable to model and utilize heterogeneous and complex data types in recommendation systems. 
However, the traditional collaborative filtering techniques such as Matrix Factorization (MF) cannot deal with the above problems in a unified manner. 
Knowledge Graph Embedding (KGE) ~\cite{wang2017knowledge,cai2018comprehensive}, 
also known as a translation-based embedding model, encodes the side-information to improve the performance of Recommender Systems (RS)  \cite{palumbo2017entity2rec,zhang2016collaborative}.
He \etal \cite{he2017translation} and Zhang \etal \cite{zhang2016collaborative} employ the KGE model to represent users, movies, and movie attributes. The graph edges represent connections between users and movies in a knowledge base \cite{he2017translation,zhang2016collaborative}. However, previous studies do not offer insights on the following challenges: (1) how to construct a user-POI graph that utilizes the user check-in data with side information, such as spatio-temporal data and semantic context information, to leverage data sparsity problem; (2) how to effectively integrate a translation-based embedding model with a traditional recommendation system to improve the quality of POI recommendation.

\vspace{0.05cm}
\noindent{\textbf{Problem definition.}} 
Given an LBSN user check-in dataset, we aim to recommend each user with  personalized top-\textit{k} POIs they may be interested in visiting.

We build upon recent advances in 
graph embedding methods and propose \textit{Graph Embedding for POI Recommendation}, a novel translation-based graph embedding approach specifically for POI recommendation, abbreviated as \textit{GERec}. 
To overcome the challenge stemming from the spatio-temporal context, GERec encodes temporal and spatial information, as well as user dynamic check-in activities in a low-dimensional latent space. 
GERec addresses the issues of user check-in data sparsity by integrating user-POI graph embedding with a combined matrix factorization framework (Fig. \ref{fg:overview}). 

\vspace{0.05cm}
\noindent{\textbf{Contributions.}} 

\renewcommand{\labelenumi}{\Roman{enumi}.}
\vspace{-2mm}
\hspace{-1cm}
\begin{enumerate}[leftmargin=1cm]
\item 
To deal with the data sparsity, we propose a novel translation-based POI recommendation model to effectively form a user-POI graph capturing the side information, such as spatial, temporal, and semantic contents.

\item We propose a spatio-temporal relation path embedding to model the temporal, spatial and semantic content information to improve the quality of POI recommendations.



\item We show our model outperforms the state-of-the-art POI recommendation techniques on two real-world LBSN datasets, Foursquare\cite{cheng2011exploring} and Gowalla\cite{cho2011friendship}. 

\end{enumerate}


\section{Related Work}
\vspace{0.05cm}
\noindent{\textbf{POI recommendations.}} 
Making personalized POI recommendations is challenging due to the user dynamic check-in activities. Existing studies on MF-based POI recommendations either focus on aggregating spatially-wise personal preference or exploring temporal influences. 
Most aggregation-based POI recommendation approaches fail to capture jointly geographical and temporal influences with the semantic context while addressing the data sparsity in an unified framework. 
In \cite{wang2015geo,yin2017spatial}, geographical locations are used to improve the performance of POI system which highlights that there is a strong correlation between user check-in activities and geographical distance. 
Geographical sparse additive generative model \cite{wang2015geo} for POI recommendations, Geo-SAGE, exploited co-occurrence patterns with contents of spatial items. 
A POI system \cite{yin2017spatial} based on deep learning from heterogeneous features and hierarchically additive representation learning proposed spatially-aware model for personal preferences. 

\vspace{0.05cm}
\noindent{\textbf{Knowledge Graph Embedding.}} 
KGE is well known for its use in recommendation systems. 
Zhang \etal \cite{zhang2016collaborative} proposed a 
collaborative KG-integrated movie recommender framework to learn the latent and visual representations. 
Palumbo \etal \cite{palumbo2017entity2rec} proposed {\em entity2rec} to capture a user-item relatedness from KGEs with the goal of generating top-$k$ item recommendations. 
Qian \etal \cite{qian2019spatiotemporal} adopts KGE to model the side information in POI recommendation system. Their work only focuses on embedding the user and POI entities, and mapping the spatio-temporal patterns as a translation matrix. Although their work explored KGEs and RS, it did not integrate graph embedding with the traditional MF model. Compared with our proposed model that combines the spatial and temporal information with semantic contents in a semantic relation embedding space, these KG embedding-based recommender models have limited expressive abilities as they model the key parameters (\eg, spatial and temporal information) as a simple matrix.  
%
Finally, noteworthy is the family of Graph Convolutional Networks with models such as GCN \cite{kwSSC}, GraphSAGE \cite{hyIRL}, adversary GCN \cite{Sun2019FisherBuresAG}, kernel-based CKN \cite{ckns} as well as generic graph embedding approaches such as DeepWalk ~\cite{paDOL} and Node2Vec~\cite{glNSF} which all have the capacity to model graph-related tasks.

\vspace{0.05cm}
\noindent{\textbf{Difference with existing works.}} 
1) To the best of our knowledge, this is the first work that investigates the joint modeling of temporal, geographical and semantic category information integrated with KG embedding  POI recommendation system; 2) A novel embedding is proposed to bridge the gaps between embedding  and traditional MF. Therefore, we propose a novel combined MF framework for dynamic user-POI preference modeling based on the learned embedding in a unified manner; 
3) In contrast to the approach \cite{xie2016learning} based on the bipartite graph (homogeneous graph), our approach uses the translation-based graph (heterogeneous graph). Moreover, approach \cite{xie2016learning} does not apply MF while our model investigates MF for generating top-$k$ proposals.  


\section{Proposed Approach}



\subsection{User-POI Graph Embedding}


A heterogeneous graph admits two or more node types which can be then embedded by a symmetric function \eg, one can use interchange \textit{user type} and \textit{POI type}  input arguments. For the best recommendation performance, we develop an effective representation for the {\em user} and {\em POI} nodes. A user $u$ and a POI $v$ represent the $head$ or $tail$ of a triplet $(head, relation, tail)$, denoted as $(u, r, v)$, where $\boldsymbol{u}, \boldsymbol{e}, \boldsymbol{v}\in\mathbb{R}^k$ are the vector representations of  $u$, $r$ and $v$.




Head-tail entity pairs usually exhibit diverse patterns in terms of relations \cite{lin2015learning}. Thus, a single relation vector cannot perform all translations between head and tail entities. For example, the relation path embedding has the diversity patterns, such as temporal, spatial and  semantic contents patterns. The relation between user (head) and POI (tail) ``user - sushi shop" exhibits many patterns: (i) Temporal pattern \ie, a user visits a POI in a certain time slot $<$user, /time slot, POI$>$; (ii) Geographical pattern \ie, a user visits a POI when she is in a particular area $<$user, /location, POI$>$; and (iii) Semantic content pattern \ie, a user visits a specific POI that is associated with a category $<$user, /category, POI$>$. In our model, we embed spatio-temporal information as a relationship connecting users and POIs. 

Take the $2$-step path as an example. In Table \ref{tb:relation_path2}, a user check-in activity (a user visits a POI) is associated with temporal and geographical patterns \ie, 
$user \stackrel{12pm}{\longrightarrow} e_{1} \xrightarrow{food court} \textit{sushi shop}$ denotes a user visiting a sushi shop (POI) at $12$pm (time slot) at food court (location). Instead of building triplets $(u, r_{t}, e_{1})$ and $(e_{1}, r_{l}, v)$ for learning the graph representation, we form a triplet $(u, r_{t} \circ r_{l}, v)$, and optimize the objective $u + (r_{t} \circ r_{l}) = v$. The composition operator $\circ $ merges the temporal and spatial relations $r_{t}$ and $r_{l}$ into the spatio-temporal relation. 
Given a relation path $r = (r_{1}, \dots, r_{n})$, we obtain the relation path embedding $\textbf{r}$ by composing multiple relations via the operator $\circ$, \ie, $\textbf{r} = \textbf{r}_{1} \circ \dots \circ \textbf{r}_{n}$. 
For the composition operator, we use
the multiplication operation.
Thus, the relation path vector is defined as 
$\textbf{r} = \textbf{r}_{1} \times \dots \times \textbf{r}_{n}$.
In our model, we embed temporal and geographical patterns, and semantic category contents into the relation path. 
For instance, $u \stackrel{r_{t}}{\longrightarrow} e_{1} \stackrel{r_{l}}{\longrightarrow} e_{2} \stackrel{r_{c}}{\longrightarrow} v$ illustrates that a user visits a POI at a certain time slot $t$ in location $l$, which has semantic category information $c$ associated with the user's current location. We define a spatio-temporal and semantic-based relation path representation $ \textbf{r}_{tlc} = \textbf{r}_{t} \circ \textbf{r}_{l} \circ \textbf{r}_{c}$, which consists of a temporal relation path $\textbf{r}_{t}$, a geographical relation path $\textbf{r}_{l}$, and a semantic relation path $\textbf{r}_{c}$. The relation $\textbf{r}_{tlc}$ is used as our default relation representation in our POI  model. In what follows, we write $\textbf{r}$ instead of $\textbf{r}_{tlc}$ for simplicity.

TransR \cite{lin2015learning,wang2017knowledge} is among the most representative translational distance models for a heterogeneous graphs. 
We apply TransR \cite{lin2015learning} to our POI recommendation model.  For each triplet, including $(u, r_{tl}, v)$ and $(u, r_{tlc}, v)$ in the graph, entities are embedded into vectors $\textbf{u}, \textbf{v} \in \mathbb{R} ^k$ and relation is embedding into $\textbf{r} \in \mathbb{R} ^d$. For each relation $r$, we set a projection matrix from the entity space to the relation space, denoted as ${\boldsymbol{M}}_{r} \in \mathbb{R}^{k \times d}$. 
TransR firstly maps entities $u$ and $v$ into the subspace of relation $r$ by using matrix $\boldsymbol{M}_r$:
\begin{equation}\label{eq:project_vector}
\textbf{u}_r = \textbf{u} \textbf{{$\boldsymbol{M}$}}_r  \quad\text{and}\quad
\textbf{v}_r = \textbf{v} \textbf{{$\boldsymbol{M}$}}_r,
\end{equation} 
and the TransR score function is defined as:
\begin{equation}\label{eq:score_func}
f_r(u,v) = \parallel \textbf{u}_r + \textbf{r} - \textbf{v}_r\parallel ^2_2.
\end{equation}

The following margin-based ranking loss  defined in \cite{lin2015learning} is used for training:
\begin{equation}
L = \sum_{(u, r, v)\in S} \sum_{(u', r, v')\in S'} max(0, f_r(u,v) + \gamma - f_r(u',v')),
\end{equation}

\noindent where $\gamma$ controls the margin between positive and negative samples, $S$ and $S'$ are the set of positive and negative triplets, respectively. The existing graphs that we construct from user-POI check-in datasets contain mostly correct triplets. 
Thus, we corrupt the correct triplet $(u, r, v) \in S$ to construct incorrect triplets $(u', r, v')$ by replacing either head or tail entities with other entities from the same group so that:
\begin{equation}
    S' = \{(h', r, t)\}\cup\{(h, r, t')\}.
\end{equation}

We note that translation-based embedding provides a generic way for extracting a useful information from a graph. However, embedding cannot be applied directly to matrix factorization models. Thus, we propose a function $g(\cdot)$ that extracts the learnt entities. 
Given an entity $u$, $v$ and a relation $r$, we obtain representation sets  $\{\textbf{e}^{r}_u\}$ and $\{\textbf{e}^{r}_v\}$, where $r$ denotes the set of relation paths, where  $\textbf{e}^{r}_u$ and $\textbf{e}^{r}_v$ represent a user $u$ and a POI $v$ with respect to the specific relation path $r$. Thus, the entity extraction is denoted as:
\begin{equation}
\{\boldsymbol{\phi}_u\} \leftarrow g(\{\textbf{e}^{r}_u\}), \quad \quad
\{\boldsymbol{\phi}_v\} \leftarrow g(\{\textbf{e}^{r}_v\}),
\end{equation}
\noindent where $\{\boldsymbol{\phi}_u\}$  and $\{\boldsymbol{\phi}_v\}$ are sets of final representations for user and POI embedding, respectively. 
The function $g(\cdot)$ prepares the embedded user and POI information to become  the entries for the matrix factorization by  
sorting the learnt user and POI embedding sets based on the distances from Eq. \eqref{eq:score_func} sorted according to the descending order. Embedded pairs that are further from each other than some $\theta$ are pruned. 
When a user connects with a POI by a relation ($\textbf{u} + \textbf{r} \approx \textbf{v}$), the smaller the score value,  the
lower distance between POI and user is.  Hence, ($\textbf{v} + \textbf{r} \approx \textbf{u}$) vice versa. Then, the sorted user and POI embedding sets are filtered according to the user's current location.  
In many cases, POIs may be outside of the user's home location and it may be not reasonable to recommend such POIs.  Thus, we set a reasonable radius \wrt the geographical location by applying a threshold $\theta_d$  to filter the learnt POIs that are too far away from user's home location. 
Following \cite{lichman2014modeling,wang2018tpm}, we assume a Gaussian distribution for user current location $l$, and  we set the user's current check-in POI $v_l$ so that $v_l \sim \mathcal{N}\left(\mu_{l},  \Sigma_{l}\right)$.

\subsection{The Combined Matrix Factorization}
We integrate the matrix factorization into our model by combining two parts: \textit{1) spatio-temporal MF} and \textit{2) User preference  MF}. The spatio-temporal MF calculates the probability that a user will visit a POI. The user preference MF evaluates user's preference \wrt a POI. 
The combined probability determines the total probability of a user $u$ visiting a POI $v$.


\subsubsection{\textbf{Spatio-temporal MF.}}
For each embedded user vector $\boldsymbol{\phi}_u$ and embedded POI vector $\boldsymbol{\phi}_v$, we apply the matrix factorization to predict a probability that a user $u$ would visit a POI $v$ based on her current location $l$ and a particular time slot $t$. Given a frequency matrix $\boldsymbol{P}' \in \mathbb{R}^{\mid \{\boldsymbol{\phi}_u\} \mid \times \mid \{\boldsymbol{\phi}_v\} \mid}$, which represents the number of check-ins of the embedded users for the embedded POIs. 
MF is performed by finding two low-rank matrices: a user specific matrix $\boldsymbol{E} \in \mathbb{R}^{K \times \mid \{\boldsymbol{\phi}_u\} \mid}$ and a POI specific matrix $\boldsymbol{O} \in \mathbb{R}^{K \times \mid \{\boldsymbol{\phi}_v\} \mid}$, where $K$ is the dimension of the latent vector that captures the corresponding user-POI preference transition. 
The probability of an embedded user $u$ based on a particular spatio-temporal relation $r_{tl}$ and embedded location $v$, is determined by:
\begin{equation}\label{eq:mf_p1}
    P'_{uv}= {\boldsymbol{E}_u}^{\!\top} {\boldsymbol{O}_v},
\end{equation}
where $\boldsymbol{E}_u$ and $\boldsymbol{O}_v$ are vectors for the user $u$ and the POI $v$ from matrices $\boldsymbol{E}$ and $\boldsymbol{O}$, respectively, while $P'_{uv}$ is a scalar frequency for $u$ and $v$.

The goal of matrix factorization is to accurately approximate the probabilities for the user frequency data:
\begin{equation}\label{eq:mf_train_func}
\underset{\boldsymbol{E},\, \boldsymbol{O}}{min} \;\;\alpha(\parallel\boldsymbol{E}\parallel^2_F + \parallel\boldsymbol{O}\parallel^2_F)\; +\! \sum_{(u,v)\in \Omega} (P'_{uv} -  {\boldsymbol{E}_u}^{\!\top} \boldsymbol{O}_v )^2 ,
\end{equation}
where $(u,v)\in \Omega$ indicates the observed frequency of user $u$ at POI $v$,  $||~\cdot~||^2_F$ is the Frobenius norm, 
and $\alpha (\parallel \boldsymbol{E} \parallel^2_F + \parallel \boldsymbol{O} \parallel^2_F)$ is a regularization term to prevent overfitting.

\subsubsection{\textbf{User preference  MF.}}
The second part of the combined MF model is to predict the user preference given a POI. Based on the user historical check-in frequency, given an observed 
frequency matrix $\boldsymbol{F}$, MF factorizes users and POIs so that $\boldsymbol{F} \approx \boldsymbol{U}^\top \boldsymbol{V}$. Then,  
scalar $P''_{uv}$ captures users' preference at a POI determined by the following equation: 
\begin{equation}\label{eq:mf_p2}
   P''_{uv}= \boldsymbol{U}^{\!\top}_u\boldsymbol{V}_v 
\end{equation}
The same objective as in Eq. (\ref{eq:mf_train_func}) is applied to accurately
approximate the probabilities for the user check-in frequencies.

\subsubsection{\textbf{Combined Matrix Factorization.}}
We propose a combined MF model that is simply a product of probabilities that 1) a user is spatio-temporally compatible with a POI and 2) the user has a preference given the POI.  
The first term  is the probability of an embedded user visiting an embedded POI given some spatio-geographic pattern, where 
$P'_{uv}$ 
is defined by Eq. (\ref{eq:mf_p1}). 
The second term is the probability of the user's preference at a POI based on her historical records, where 
$P''_{uv}$ 
is defined in Eq. (\ref{eq:mf_p2}). 
The combined model is denoted as:
\begin{equation}
    P_{uv} = 
    P'_{uv}\!\cdot\! P''_{uv}.
\end{equation}


\section{Experiments}


\subsection{Experimental Configuration}

  
\textbf{Datasets.} We adopt two popular large-scale LBSN datasets: Foursquare \cite{cheng2011exploring} and Gowalla \cite{cho2011friendship}. 
The experimental results for our approach and the baselines are compared in the same testbed. 
We selected the Foursquare dataset from Sep 2010 to Jan 2011 which contains 1,434,668 users’ check-in activities in the USA. 
The Foursquare geographical area is divided into a set of $5846$ locations/regions according to administrative divisions. There are $114,508$ user entities and $62,462$ POI entities connected with $46,768$ spatio-temporal relations.  
For Gowalla, another graph is built from $107,092$ user entities and $1,280,969$ POI entities connected with $1,633$ relations. 
We apply $k$-means  \cite{yin2016adapting,qian2019spatiotemporal} to form $200$ region clusters for Gowalla geographical area.

\noindent{\textbf{Baselines.}} 
Two of the baseline models are translation-based models that are highly related work in RS  \cite{he2017translation,qian2019spatiotemporal}. \textbf{PMF} \cite{mnih2008probabilistic} is a classic probabilistic matrix factorization model that explicitly factorizes the rating matrix into two low-rank matrices. 
\textbf{GeoMF} \cite{lian2014geomf} is a weighted matrix factorization model for  POI recommendations. 
\textbf{Rank-GeoFM} \cite{li2015rank} is a ranking-based geographical factorization model in which the check-in frequency
characterizes users' visiting preference, and the factorization is learnt by ranking  POIs. 
\textbf{GeoSoCa} model \cite{zhang2015geosoca} extends the kernel density estimation by applying an adaptive bandwidth learnt from the user check-in data. 
\textbf{ST-LDA} \cite{yin2016adapting} is a latent class probabilistic generative Spatio-Temporal LDA (Latent Dirichlet Allocation) model, which learns the region-dependent personal interests according to the contents of the checked-in POIs at each region. 
\textbf{TransRec} is the translation-based recommendation approach proposed in \cite{he2017translation}, which embeds items into a translation space and models users via a translation vector. 
Note that our proposed method is different from TransRec as we select both users and POIs as entities, and learn the embedding representation for a different type of knowledge as well as the spatio-temporal relationships. 
\textbf{STA}
\cite{qian2019spatiotemporal} is a spatio-temporal context-aware and translation-based POI recommendation model. However, this solution does not consider the semantic relation embedding of spatial, temporal and category content information, and thus is incapable of leveraging the user-POI graph structure.

\noindent{\textbf{Evaluation Metrics.}}  Following \cite{lian2014geomf,li2015rank,zhang2015geosoca}, we deploy the following evaluation methodology. 
The user-POI graph is built from historical user check-in activities in the training set. The spatio-temporal relations in the user-POI graph are composed based on each user's current time slot and the area from the given query $q = (u, l, t)$. 
We divide the time slot to different hour lengths $(1, 2, 4, 8, 12, 24)$.
The user's current standing  area before visiting $v$ is selected for her location $l$. 
In the experiment, we first calculate the frequency for each user visiting ground-truth POIs. 
We use the $80\%$ as the cut-off point so that check-ins before a particular date are used for training. The rest check-in data generated after this date is chosen for testing. We form a top-$k$ recommendation list from the top $k$ POI recommendations. We deploy  measurement metrics such as Precision$@k$ ($Prec@k$), Recall$@k$ ($Rec@k$) and F1-score$@k$ ($F@k$): 
$Prec@k = \frac{1}{M} \sum^M_{u=1} \frac{|V_u(k) \cap V_u|}{k}$ and $Rec@k = \frac{1}{M} \sum^M_{u=1} \frac{|V_u(k) \cap V_u|}{|V_u|}$.



\subsection{Main evaluations}

\begin{figure*}[tb]
\begin{center}
\centering
  \subcaptionbox{Prec@K on Foursquare\label{fig3:ba}}{\includegraphics[width=1.58in]{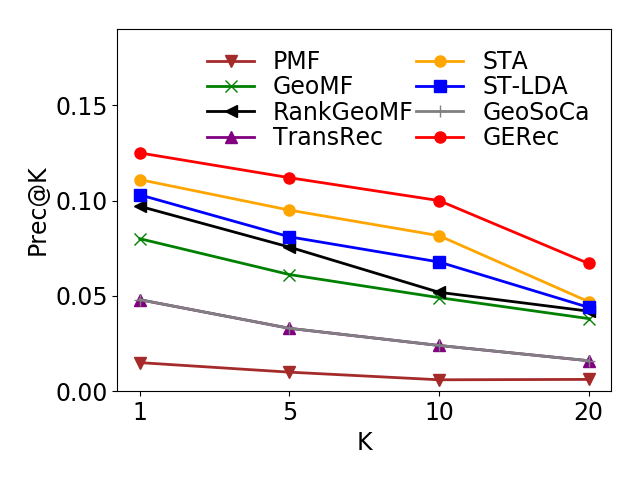}}
  \subcaptionbox{Rec@K on Foursquare\label{fig3:bb}}{\includegraphics[width=1.58in]{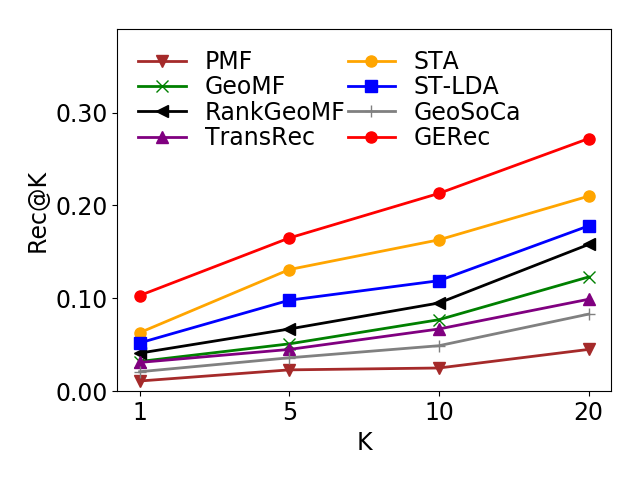}} 
  \subcaptionbox{F1@K on Foursquare\label{fig3:bc}}{\includegraphics[width=1.58in]{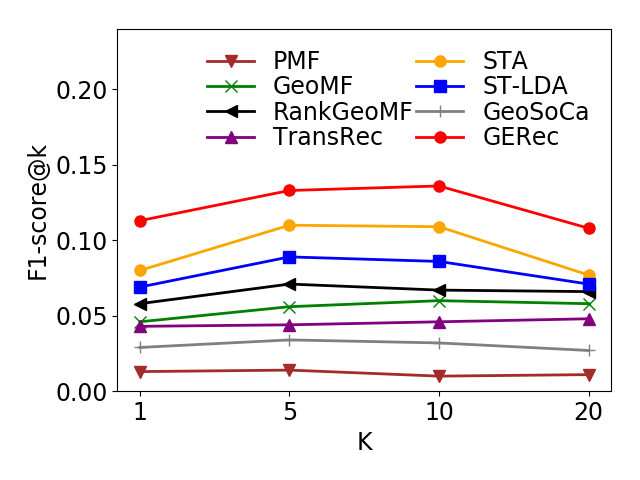}}
  \\
  \subcaptionbox{Prec@K on Gowalla\label{fig3:bd}}{\includegraphics[width=1.575in]{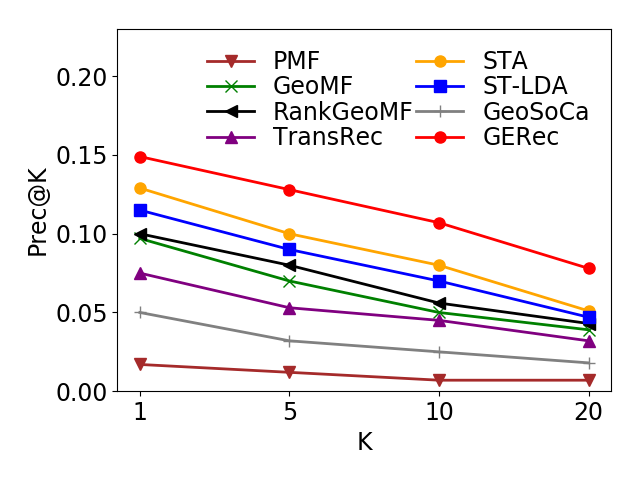}}
  \subcaptionbox{Rec@K on Gowalla \label{fig3:be}}{\includegraphics[width=1.575in]{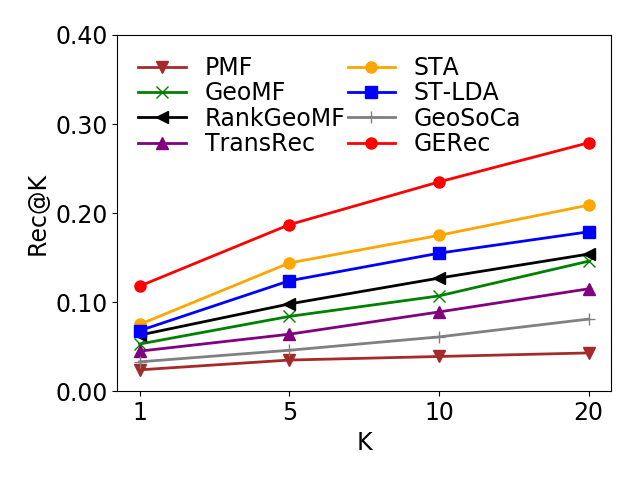}}
  \subcaptionbox{F1@K on Gowalla\label{fig3:bf}}{\includegraphics[width=1.575in]{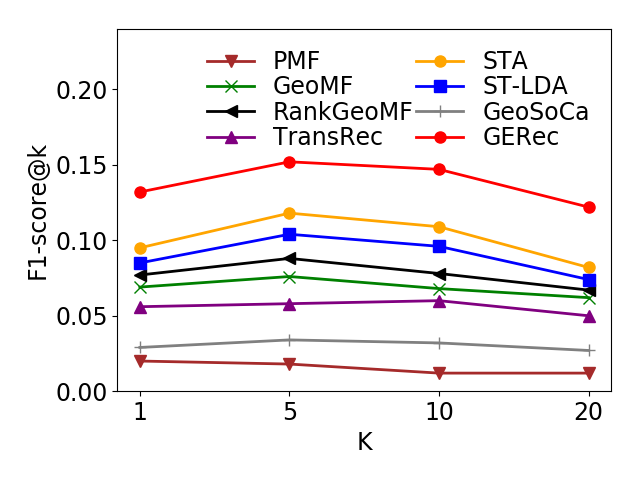}}
  \end{center}
    \caption{Baseline comparisons.}
  \label{fg:baseline}
\end{figure*}

Following \cite{lin2015learning,lin2015modeling}, 
for translational distance model TransR, we set the learning rate $\lambda = 0.001$, the margin $\gamma = 1$, the dimensions of entity embedding and relation embedding $d = 100$, the batch size $B = 120$. We traverse all the training triplets for 1000 rounds on both Foursquare and Gowalla datasets. Fig.~\ref{fg:baseline} reports the performance of the POI recommendation models on Foursquare and Gowalla datasets, respectively. We present the performance for $k = \{1, 5, 10, 20\}$. 
Fig.~\ref{fg:baseline} presents the results of  algorithms in terms of $Prec@k$, $Rec@k$ and $F1@k$ on Foursquare and Gowalla datasets. 
The figure show that the proposed GERec model outperforms all baseline models significantly for all metrics at different $k$ values. 
Specifically, 
when comparing with the traditional MF models, GERec outperforms the Rank-GeoMF, which is the MF baseline with the best performance, by 50\% and 47\% in F1-score@10 on Foursquare and Gowalla, respectively. When comparing with translation-based models, our proposed model also improves the POI recommendation performance significantly. GERec outperforms STA, by 20\% and 25\% in F1-score@10 on both datasets. This demonstrates the capability of our graph-based  GERec model to generate high quality POI recommendations. 
Although GeoSoCa exploits social influences, geographical locations and user interests, the simple kernel density estimation results in the poor performance. This validates the effectiveness of our GERec solution, especially our proposed step which exploits and integrates the user-POI interactions and spatio-temporal patterns to tackle the sparsity in the user-POI check-in data. 
The learned embeddings are well integrated into the combined matrix factorization model.  
Thus, GERec achieves the best performance among all compared baseline models. 
The user-POI graph constructed from Gowalla dataset has fewer relation edges than the Foursquare graph, as Gowalla relation patterns $r_{tl}$ have fewer regions in the relation paths than Foursquare.



\subsection{Impact of data sparsity}
\begin{figure*}[tp]
\begin{center}
\centering  
  \subcaptionbox{Prec@10 on Foursquare\label{fig3:cb}}{\includegraphics[width=1.575in]{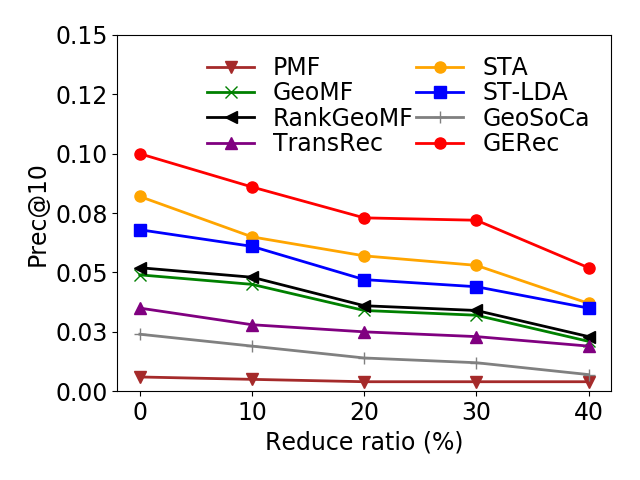}}
  \subcaptionbox{Rec@10 on Foursquare\label{fig3:cd}}{\includegraphics[width=1.575in]{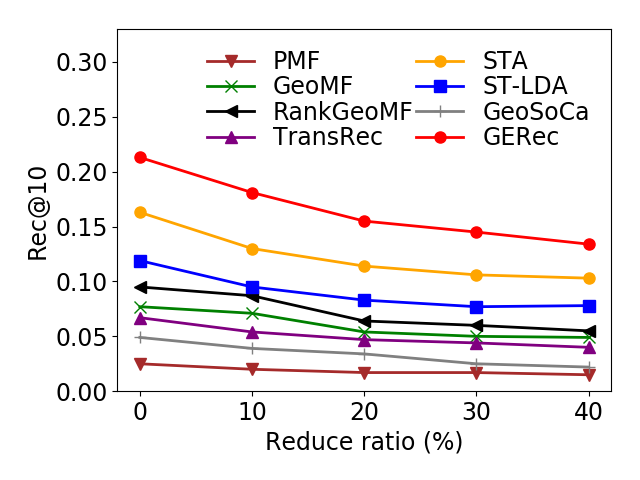}}
  \subcaptionbox{F1@10 on Foursquare \label{fig3:aa}}{\includegraphics[width=1.575in]{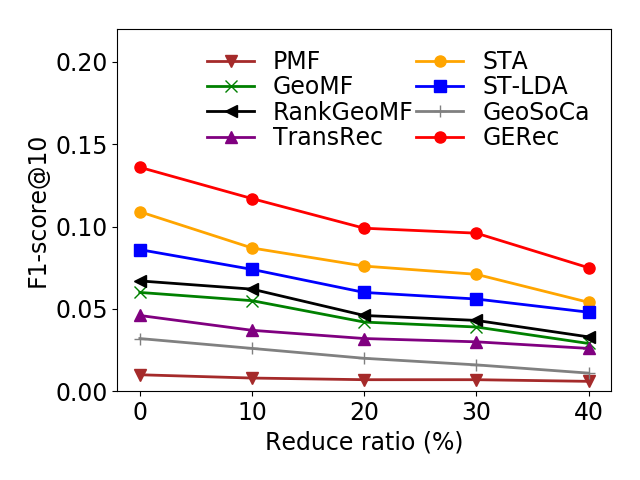}}
  \\
  \subcaptionbox{Prec@10 on Gowalla \label{fig3:ab}}{\includegraphics[width=1.575in]{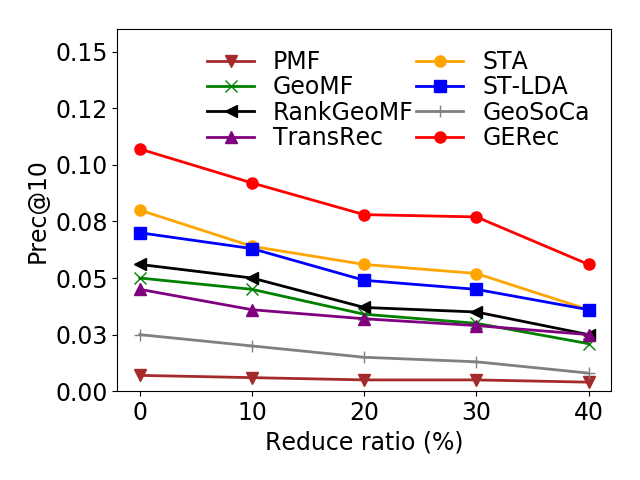}}
  \subcaptionbox{Rec@10 on Gowalla \label{fig3:ac}}{\includegraphics[width=1.575in]{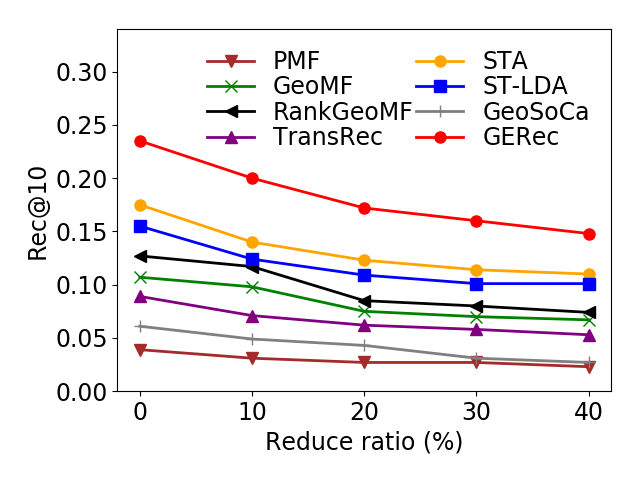}}
  \subcaptionbox{F1@10 on Gowalla \label{fig3:ad}}{\includegraphics[width=1.575in]{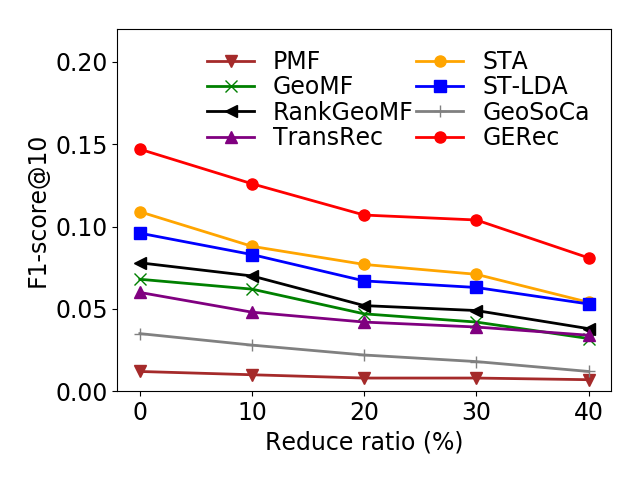}}
\end{center}
\caption{Sensitivity to data sparsity.}
\label{fg:sparsity}
\vspace*{-0.18cm}
\end{figure*}

In Fig.~\ref{fg:sparsity}, we conduct extensive experiments to evaluate the performance of the models under the data sparsity. Specifically, we create multiple datasets with various sparsity levels by reducing the amount of training data randomly by 10\%, 20\%, 30\%, and 40\% of the total amount of data (before the cut-off date), and at the same time keeping the test data the same. 
The results on both Foursquare and Gowalla are shown in the Fig.~\ref{fg:sparsity}. Specifically, the $0$ in the horizontal axis presents the experiment result without reducing the training data. We observe that the Precision@$k$ and Recall@$k$ value decrease for all baseline models.
For example, the performance of these models in terms of Prec@10 decreases at least 40\% on Foursquare when reducing 40\% of the training data. Results of the ST-LDA  drop significantly compared with the other baseline models with 37\% drop in Rec@10 on both Foursquare and Gowalla, which indicates that the LDA-based model is sensitive to the sparse data. The PMF does not change much, however, it remains the least accurate result. The Rec@10 values of RankGeoMF, TransRec, and STA show a 42\%, 41\%, and 37\% drop, respectively. The GERec drops by only 34\%, which illustrates that our proposed model is more stable and robust under sparsity than the baselines.

\subsection{Impact of time slot and dimensionality}


There are two parameters in the proposed GERec model: the time slot $h$ and the embedding dimension $d$. Below, we investigate the effect of these two parameters. 
Table \ref{tb:timeslot} shows the impact of the length of time slot on Precision@$k$ and Recall@$k$. The length of the time slot affects the quality of POI recommendations. When the length of time slot changes, the relation paths change and the entire  graph needs to be computed again. We split day activities into different lengths. The parameter $h$ denotes the length of each time slot in hours. The larger length of time slot the less the time influence on recommendation results. We report the top-$k$ recommendation precision and recall for each time slot on the Foursquare and Gowalla datasets. From the experimental results we observe that the POI recommendation accuracy improves when the time slot length increases. The recommendation accuracy reaches a peak point for $8$ hours long time slot. Then, it starts decreasing as the time slot keeps increasing. The reason for the improved accuracy is that the larger time length, the denser the data. Hence, there are more user check-in records at each time slot for generating recommendations. However, the recommendation accuracy decreases as the length of the time slot reaches $8$ hours.  This is because when the length of the time slot is large enough, it may reduce the influence of temporal pattern. 
Moreover, we study the impact of varying dimension $d$ of the relation embedding by setting it to $\{70, 80, 90, 100, 120\}$ (Table \ref{tb:dim}). The best parameter is determined according to the mean rank in the test set. The accuracy rate increases gradually when the dimension increases. Specifically, the accuracy keeps increasing until the dimension reaches $100$, then it remains stable. 
\begin{table*}[tp]
\centering
\caption{Impact of the time slot length $h$.}
\label{tb:timeslot}
\centering 
\scalebox{0.75}{
\begin{tabular}{cccc|ccc|ccc|ccc}
\hline
\multirow{2}{*}{Hours} &
\multicolumn{6}{c}{Foursquare} & \multicolumn{6}{c}{Gowalla} \Tstrut\Bstrut \\
\cline{2-13}
& Prec@1 & Prec@10 & Prec@20 & Rec@1 & Rec@10 & Rec@20
& Prec@1 & Prec@10 & Prec@20 & Rec@1 & Rec@10 & Rec@20 \Tstrut\Bstrut  \\
\hline
1 & 0.075 & 0.061 & 0.041 & 0.062 & 0.128 &	0.163
& 0.089 & 0.064 & 0.047 & 0.071 & 0.141 & 0.169
\Tstrut\Bstrut \\
2 & 0.100 & 0.083 & 0.054 & 0.082 & 0.170 & 0.218
& 0.119 & 0.086 & 0.062 & 0.094 & 0.150 & 0.225
\Bstrut \\
4 &	0.113 & 0.090 & 0.060 & 0.092 & 0.192 & 0.245
& 0.134 & 0.096 & 0.070 & 0.106 & 0.168 & 0.253
\Bstrut \\
8 &	\textbf{0.125} & \textbf{0.100} & \textbf{0.067} & \textbf{0.103} & \textbf{0.213} & \textbf{0.272}
& \textbf{0.149} & \textbf{0.107} & \textbf{0.078} & \textbf{0.118} & \textbf{0.187} & \textbf{0.281}
\Bstrut \\
12 & 0.119 & 0.095 & 0.064 & 0.098 & 0.202 & 0.258
& 0.142 & 0.102 & 0.074 & 0.112 & 0.178 & 0.267
\Bstrut \\
24 & 0.115 & 0.092 & 0.062 & 0.094 & 0.196 & 0.250
& 0.137 & 0.098 & 0.072 & 0.108 & 0.172 & 0.259
\Bstrut \\
\hline 
\end{tabular}}
\vspace{-0.1cm}
\end{table*}

\begin{table*}[tp]
\centering
\vspace*{-0.3cm}
\caption{Impact of dimensionality $d$.}
\label{tb:dim}
\centering 
\small
\scalebox{0.75}{
\begin{tabular}{cccc|ccc|ccc|ccc}
\hline
\multirow{2}{*}{$d$} &
\multicolumn{6}{c}{Foursquare} & \multicolumn{6}{c}{Gowalla}  \Tstrut\Bstrut \\
\cline{2-13}
& Prec@1 & Prec@10 & Prec@20 & Rec@1 & Rec@10 & Rec@20
& Prec@1 & Prec@10 & Prec@20 & Rec@1 & Rec@10 & Rec@20
\Tstrut\Bstrut  \\
\hline
70 & 0.121 & 0.097 & 0.065 & 0.099 & 0.206 & 0.262
& 0.143 & 0.123 & 0.075 & 0.113 & 0.226 & 0.270
\Tstrut\Bstrut \\
80 & 0.123 & 0.098 & 0.066 & 0.101 & 0.209 & 0.267
& 0.146 & 0.125 & 0.076 & 0.115 & 0.183 & 0.275
\Bstrut \\
90 & 0.124 & 0.099 & 0.066 & 0.102 & 0.211 & 0.269
& 0.148 & 0.127 & 0.077 & 0.117 & 0.185 & 0.278
\Bstrut \\
100 & \textbf{0.125} & \textbf{0.100} & \textbf{0.067} & \textbf{0.103} & \textbf{0.213} & \textbf{0.272}
& \textbf{0.149} & \textbf{0.128} & \textbf{0.078} & \textbf{0.118} & \textbf{0.187} & \textbf{0.281}
\Bstrut \\
110 & 0.126 & 0.100 & 0.067 & 0.103 & 0.214 & 0.273
& 0.150 & 0.129 & 0.078 & 0.118 & 0.188 & 0.282
\Bstrut \\
120 & 0.126 & 0.101 & 0.067 & 0.103 & 0.214 & 0.274
& 0.150 & 0.129 & 0.079 & 0.119 & 0.188 & 0.283
\Bstrut \\
\hline 
\end{tabular}}
\vspace{-0.1cm}
\end{table*}

\section{Conclusions}
In this paper, we propose a novel translation-based POI recommendation model, which can effectively construct a user-POI graph and model the side information. 
To address time and geographical reasoning, we propose spatio-temporal relation path embedding to model the temporal, spatial and semantic contents to leverage the user-POI interaction and improve the quality of user embedding. 
To overcome the sparsity of the user-POI interaction data, we develop an embedding function which bridges gaps between the translation-based embedding model and traditional MF-based model.  
The user-POI graph is integrated with a combined MF model to improve the quality of POI recommendations.

\section*{Acknowledgments}
We acknowledge the support of Australian Research Council Discovery \textit{DP190101485}, Alexander von Humboldt Foundation, and CSIRO Data61 Scholarship program.


\bibliographystyle{splncs04}
\bibliography{main}

\end{document}